%% file: root.tex
\documentclass[letterpaper, 10 pt, conference]{Classes/ieeeconf}  
\IEEEoverridecommandlockouts 
\usepackage{Packages/decar-common}
\usepackage{Packages/decar-dynamics}
\usepackage[backend=bibtex,bibstyle=ieee,citestyle=numeric-comp,doi=false,isbn=false]{biblatex}
\usepackage{graphicx}
\usepackage{lipsum}
\usepackage{derivative}
\usepackage{soul}

\usepackage{tikz, pgfplots}
\usepackage[europeanresistors]{circuitikz}
\usetikzlibrary{positioning}

\bibliography{refs}
\graphicspath{{figs/}}

\title{\LARGE \bf
   Koopman Representation of Nonlinear Virtual Environments in Kinesthetic Haptic Systems
}

\author{Yanting Zhou, Jozsef Kövecses, and James Richard Forbes}

\begin{document}

%
%
%
%
%
%
%
\def \myYear {2026}


\begin{figure*}[t]

\thispagestyle{empty}
\begin{center}
\begin{minipage}{6in}
\centering
This paper has been accepted for publication in \textit{IEEE International Conference on Robotics and Automation}. 
\vspace{1em}

This is the author's version of an article that will published in this conference. Changes will be made to this version by the publisher prior to publication.
\vspace{2em}


\vspace{2em}


\vspace{15cm}
\copyright \myYear \hspace{4pt}IEEE. Personal use of this material is permitted. Permission from IEEE must be obtained for all other uses, in any current or future media, including reprinting/republishing this material for advertising or promotional purposes, creating new collective works, for resale or redistribution to servers or lists, or reuse of any copyrighted component of this work in other works.

\end{minipage}
\end{center}
\end{figure*}
\newpage
\clearpage
\pagenumbering{arabic} 

    \maketitle
    \thispagestyle{empty}
    \pagestyle{empty}
    
    \input{Sections/abstract.tex}
    \input{Sections/introduction.tex}
    \input{Sections/methods.tex}
    \input{Sections/Experiment.tex}

    \input{Sections/results.tex}
    \input{Sections/conclusions.tex}
    \input{Sections/appendix.tex}

    \input{Sections/acknowledgement.tex}

    \printbibliography
\end{document}

%% file: Sections/abstract.tex
\begin{abstract}
	Rendering haptic feedback with nonlinear virtual environments (VEs) is important in many applications that require highly accurate force feedback. This paper considers the use of the Koopman operator to represent a nonlinear VE interacting with a haptic system. Simulation and experimental results demonstrated that the proposed method provides an effective representation of the nonlinear dynamics of a Duffing-oscillator VE. A multi-user study further confirmed this conclusion. In addition, a closed-loop (CL) stability analysis is performed leveraging the Koopman representation of the nonlinear VE to access stability of the overall haptic system. This alternative way of representing nonlinear VEs enables a convenient CL stability analysis that is less conservative than traditional passivity-based methods. Since a linear combination of all lifted states is used to represent the nonlinearity, such representation is also more robust to uncertainties in the modeling of the haptic device than a traditional nonlinear model.
\end{abstract}

%% file: Sections/introduction.tex
\section{Introduction}
The sense of touch plays a fundamental role in human perception and interaction with the physical world. By rendering force feedback, kinesthetic haptic devices allow humans to perceive touch when interacting with virtual environments (VEs) \cite{hannaford2016haptics_roboticHandbook}. Over the past several decades, haptic technology has been applied across a wide range of domains, such as surgical training, pilot training, medical robotics, and rehabilitation. Accurate and stable force rendering is essential in these applications to ensure safe and effective operations. However, many of these applications rely on accurate simulations of nonlinear VEs. As demonstrated in \cite{misra2007force}, a nonlinear tissue model provides a significantly different force feedback than a linear one, capturing critical perceptions in surgical operations on human tissue. This emphasizes the necessity of using nonlinear VEs for realistic haptic feedback in surgical simulation and all other applications that demand high accuracy in haptic feedback. However, simulating nonlinear VEs introduces challenges in VE modeling, as well as in the stability analysis of the overall haptic system.

Approaches have been proposed to address these challenges. A straightforward solution for simulating nonlinear VEs is to employ nonlinear dynamic models. However, this complicates stability analysis, and complex nonlinear dynamics slow down the VE integration \cite{basdogan2002haptic}. A neural-network (NN) based control method has been proposed to stabilize the overall haptic system by canceling the lumped dynamics effect of the whole system, without needing to model the VE dynamics \cite{esfandiari2017robust_NNbased_control}. However, this is computationally demanding since the weights of the NN must be updated online during the operation. A physics-driven NN-based simulation method has also been proposed in \cite{de2011physics-driven_NN_deformable_obj} for nonlinear deformable VE, but its application is limited to geometric and material nonlinearities. 

Strategies for assessing the stability of general haptic systems have been extensively studied, among which the theory of passive systems is commonly used. Passivity-based stability approaches often rely on an enforcement of passivity of the haptic system \cite{Colgate1994_sampled_data_sys, Hannaford_2002_POPC, Jee_Hwan_Ryu_2005_energy_reference_POPC}. However, such an approach is often too conservative, sacrificing transparency and fidelity of the haptic system \cite{adams1999stable}. To address the issues associated with passivity-based approaches, some studies put the haptic system in a port-network framework and examine Llewellyn’s absolute stability criteria \cite{adams1999stable, jazayeri2012revisiting_llewellyn}. However, the criteria developed in \cite{adams1999stable} requires the VE to be passive, while the criteria developed in \cite{jazayeri2012revisiting_llewellyn}, although relaxing the passivity constraint, is derived for linear time-invariant (LTI) VEs. Therefore, general stability criteria of haptic systems are either overly conservative or often require an LTI VE assumption. 
There also exist studies that guarantee stability specifically for haptic systems with nonlinear VEs. However, they also often rely on passivity and provide only sufficient but not necessary conditions, which is still conservative \cite{miller1999passive_nonlinVE, miller2000guaranteed_nonlinearVE}.

The principal contributions of this work are
\begin{itemize}
    \item the use of the Koopman operator framework to learn an alternative representation of the nonlinear VE for the purpose of haptic rendering, and 
    \item a stability analysis that leverages the Koopman representation that is linear and thus less complex than alternative approaches.
\end{itemize}
The key advantage of employing the Koopman operator is its ability to represent nonlinear dynamics by a linear model in a higher-dimensional space, thereby making well-established tools from linear system theory directly applicable to the analysis and control of nonlinear systems \cite{brunton2021modern}. In this work, a Duffing-oscillator VE will be used as an illustrative example in a haptic application. The Duffing oscillator is chosen since it is a widely studied benchmark system in nonlinear dynamics \cite{sastry2013nonlinear}, where the cubic stiffness term introduces essential nonlinearity that is representative for various physical systems \cite{miller2005structural}.






%% file: Sections/methods.tex
\section{Background}


This section outlines the theoretical background related to the proposed framework, including the Koopman operator theory for representing nonlinear VEs and the formulation a closed-loop (CL) stability analysis of the haptic system with a Koopman-model VE. Lyapunov’s direct method is also reviewed as a point of comparison. 

\subsection{Koopman Operator Theory with Inputs}
The Koopman operator has attracted much attention recently \cite{mezic2023operator, budivsic2012applied, mauroy2020koopman, colbrook2024limits, brunton2022data, proctor2018generalizing}. The following formulations are mainly adopted from \cite{Dahdah_2024_CL_koopman, korda2018linear, kutz2016dynamic, proctor2018generalizing}.
 Consider some discrete-time (DT) nonlinear dynamics with input $\mbf{u}_k \in {\rnums}^{n \times 1}$,
\begin{equation}
    \mbf{x}_{k+1} = \mbf{f}(\mbf{x}_k, \mbf{u}_k),
\end{equation}
where $\mbf{x}_k \in {\rnums}^{m \times 1}$ is the state vector at time $t_k$. A \textit{lifting function} ${\psi} : {\rnums}^{m \times 1} \times {\rnums}^{n \times 1} \to \rnums$ is any function that returns a scalar. The \textit{Koopman operator}, $\mc{U}$, is a linear operator that advances the lifting function by one time step. Therefore,
\begin{equation}
\label{eqn: lifting_function_advancing}
    \psi(\mbf{x}_{k+1}) = (\mc{U} \psi)(\mbf{x}_k, \mbf{u}_k),
\end{equation}
where $(\mc{U} \psi)(\cdot) = (\psi \circ \mbf{f})(\cdot)$ \cite{kutz2016dynamic}, and 
\begin{equation}
\label{eqn: koopman_with_input}
    (\mc{U} \psi)(\mbf{x}_k, \mbf{u}_k) = \psi(\mbf{f}(\mbf{x}_k, \mbf{u}_k), \star),
\end{equation}
where $\star=\mbf{u}_k$ if the input has state-dependent dynamics, or $\star = \mbf{0}$ if the input has no dynamics.

Stacking all the lifting functions together forms a \textit{vector-valued lifting function}, $\mbs{\psi} : {\rnums}^{m \times 1} \times {\rnums}^{n \times 1} \to \rnums^{p \times 1}$, which can be partitioned as
\begin{equation}
    \mbs{\psi}(\mbf{x}_k, \mbf{u}_k) = 
\begin{bmatrix}
\mbs{\vartheta}(\mbf{x}_k) \\
\mbs{\upsilon}(\mbf{x}_k, \mbf{u}_k)
\end{bmatrix},
\end{equation}
where $\mbs{\vartheta}: {\rnums}^{m \times 1} \to {\rnums}^{p_\vartheta \times 1}$, $\mbs{\upsilon}: {\rnums}^{m \times 1} \times {\rnums}^{n \times 1} \to {\rnums}^{p_\upsilon \times 1}$, and $p_\vartheta+p_\upsilon=p$. For convenient extraction of the original states, the state-dependent vector-valued lifting function often contains the original states $\mbf{x}_k$ as the first $m$ terms \cite{Dahdah_2024_CL_koopman, kutz2016dynamic},
\begin{equation}
\label{eqn: vector-valued lifting function}
    \mbs{\vartheta}(\mbf{x}_k) =
    \bbm
        \mbf{x}_k \\
        \vartheta_m(\mbf{x}_k) \\
        \vartheta_{m+1}(\mbf{x}_k) \\
        \vdots \\
        \vartheta_{p_\vartheta-1}(\mbf{x}_k)
    \ebm.
\end{equation}

With an exogenous input, (\ref{eqn: koopman_with_input}) is approximated by 
\begin{equation}
\label{eqn: koopman_matrix_approx_with_input}
    \mbs{\vartheta}(\mbf{x}_{k+1}) = \mbf{U} \mbs{\psi}(\mbf{x}_k, \mbf{u}_k) + \mbf{r}_k,
\end{equation}
where $\mbf{U} = \begin{bmatrix} \mbf{A} & \mbf{B} \end{bmatrix}$ is the Koopman matrix, and $\mbf{r}_k$ is the residual error of the approximation \cite{kutz2016dynamic}. Equation (\ref{eqn: koopman_matrix_approx_with_input}) can be expanded to a familiar linear state-space form in the lifted space,
\begin{equation}
    \label{eqn: ss_koopman}
    \mbs{\vartheta}(\mbf{x}_{k+1}) = \mbf{A} \mbs{\vartheta}(\mbf{x}_k) + \mbf{B} \mbs{\upsilon}(\mbf{x}_k, \mbf{u}_k) + \mbf{r}_k.
\end{equation}

The Koopman matrix $\mbf{U}$ is approximated by minimizing the residual in (\ref{eqn: koopman_matrix_approx_with_input}). Consider a dataset $\mc{D} = \left\{ \mbf{x}_k, \mbf{u}_k \right\}_{k=0}^{q}$ and the corresponding snapshots of the lifted states
\begin{equation}
    \mbs{\Psi} = \begin{bmatrix} \mbs{\psi}_0 & \mbs{\psi}_1 & \cdots & \mbs{\psi}_{q-1} \end{bmatrix} \in \rnums^{p \times q},
\end{equation}
\begin{equation}
    \mbs{\Theta}_{+} = \begin{bmatrix} \mbs{\vartheta}_1 & \mbs{\vartheta}_2 & \cdots & \mbs{\vartheta}_q \end{bmatrix} \in \rnums^{p_\vartheta \times q},
\end{equation}
where $\mbs{\psi}_k = \mbs{\psi}(\mbf{x}_k, \mbf{u}_k)$ and $\mbs{\vartheta}_k = \mbs{\vartheta}(\mbf{x}_k)
$. Then, considering (\ref{eqn: koopman_matrix_approx_with_input}), minimizing the residual is to minimize the least-square error
\begin{equation}
\label{eqn: Koopman_cost_function}
    J(\mbf{U}) = \f{1}{q}\left\| \mbs{\Theta}_{+} - \mbf{U} \mbs{\Psi} \right\|_\frob^2.
\end{equation}

The theoretical solution to (\ref{eqn: Koopman_cost_function}) is $\mbf{U} = \mbs{\Theta}_{+} \mbs{\Psi}^{\dagger}$, where $\mbs{\Psi}^{\dagger}$ is the Moore-Penrose pseudoinverse of $\mbs{\Psi}$.
However, $\mbs{\Psi}^{\dagger}$ can be numerically ill-conditioned when $q\gg p$ or $q\ll p$. This issue can be addressed using the Dynamic Mode Decomposition (DMD) for $q\ll p$ and Extended Dynamic Mode Decomposition (EDMD) algorithms for $q\gg p$ \cite{Dahdah_2022_sysNorm_regularization, kutz2016dynamic, williams2015data}. The complete Koopman pipeline including the DMD and EDMD algorithms can be implemented conveniently using the Python package, \texttt{pykoop} \cite{pykoop}.

It can be observed from (\ref{eqn: ss_koopman}) that the dynamics are linear in the lifted space. This facilitates the use of well-established linear tools for tasks such as stability analysis.


\subsection{Closed-Loop Stability Analysis}
\label{sec: Closed-Loop Stability Analysis}
\begin{figure}[h]
    \centering
\begin{circuitikz}[american voltages, european]
  \node[draw,minimum width=2cm,minimum height=1.4cm,align=center] (hd) {Haptic Device\\$x_\mathrm{d}, \; \dot{x}_\mathrm{d}$};
  \node[draw,minimum width=2cm,minimum height=1.4cm,align=center,right=2.6cm of hd] (ve) {VE\\$x_\mathrm{v}, \; \dot{x}_\mathrm{v}$};

  \coordinate (L) at (hd.east);
  \coordinate (R) at (ve.west);

   \draw ($(L)+(0,0.45)$) to[spring,l^=$k_{\mathrm{c}}$] ($(R)+(0,0.45)$);
  \draw ($(L)+(0,-0.45)$) to[damper,l_=$b_{\mathrm{c}}$] ($(R)+(0,-0.45)$);
\end{circuitikz}
\caption{Illustration of haptic system with virtual coupling.}
\label{fig: haptic_block}
\end{figure}
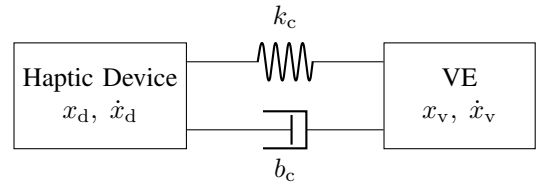

Since the example VE considered in this paper is a 1-degree-of-freedom (DoF) Duffing oscillator, for simplicity in modeling and analysis, consider a 1-DoF rigid-body model for the haptic device,
\begin{equation}
    \label{eqn: device_dynamics}
    m_{\mathrm{d}}\ddot{x}_{\mathrm{d}}(t)+b_{\mathrm{d}}\dot{x}_{\mathrm{d}}(t)=F_{\mathrm{c}}(t),
\end{equation}
where the subscript ``d" indicates variables and parameters associated with the device, $m_{\mathrm{d}}$ is the effective mass at the end-effector (EE), $b_{\mathrm{d}}$ is the damping coefficient, $x_{\mathrm{d}}(t)$ is the position, and $F_{\mathrm{c}}(t)$ is the coupling force. Consider a virtual coupling that represents the interaction between the device and the VE with virtual stiffess $k_{\mathrm{c}}$ and virtual damping $b_{\mathrm{c}}$. The coupling force that is transmitted to the device is
\begin{equation}
\label{eqn: coupling_force}
    F_{\mathrm{c}}(t)=k_{\mathrm{c}}(x_{\mathrm{v}}(t)-x_{\mathrm{d}}(t))+b_{\mathrm{c}}(\dot{x}_{\mathrm{v}}(t)-\dot{x}_{\mathrm{d}}(t)),
\end{equation} 
where $x_{\mathrm{v}}(t)$ is the VE position and $\dot{x}_{\mathrm{v}}(t)$ is the VE velocity. The setup is illustrated in Figure~\ref{fig: haptic_block}.




Let $z_1(t)=x_{\mathrm{d}}(t)$ and $z_2(t)=\dot{x}_{\mathrm{d}}(t)$ be the states of the device. Let $x_1(t)=x_{\mathrm{v}}(t)$ and $x_2(t)=\dot{x}_{\mathrm{v}}(t)$ be the states of the VE. The state-space form of (\ref{eqn: device_dynamics}) is

\begin{equation}
    \label{eqn: device_ss}
    \underbrace{\bbm\dot{z}_1(t)\\ \dot{z}_2(t)\ebm}_{\mbfdot{z}(t)}
=
\underbrace{\begin{bmatrix}
0 & 1\\
-\frac{k_{\mathrm{c}}}{m_{\mathrm{d}}} & -\frac{b_{\mathrm{d}} + b_{\mathrm{c}}}{m_{\mathrm{d}}}
\end{bmatrix}}_{\mbf{A}_{\mathrm{d,\,cont}}}
\underbrace{\begin{bmatrix}z_1(t)\\ z_2(t)\end{bmatrix}}_{\mbf{z(t)}}
+
\underbrace{\begin{bmatrix}
0 & 0\\
\frac{k_{\mathrm{c}}}{m_{\mathrm{d}}} & \frac{b_{\mathrm{d}}}{m_{\mathrm{d}}}
\end{bmatrix}}_{\mbf{B}_{\mathrm{d,\,cont}}}
\underbrace{\begin{bmatrix}x_1(t)\\ x_2(t)\end{bmatrix}}_{\mbf{x}(t)}.
\end{equation}

The DT state-space form of the device dynamics is
\begin{equation}
    \label{eqn: ss_device_DT}
    \mbf{z}_{k+1}
=
\underbrace{e^{\mbf{A}_{\mathrm{d,\,cont}}\Delta T}}_{\mbf{A}_{\mathrm{d}}}
\mbf{z}_{k}
+
\underbrace{\left( \int_{0}^{\Delta T} e^{\mbf{A}_{\mathrm{d,\,cont}}(\Delta T-\tau)}\,\mbf{B}_{\mathrm{d,\,cont}}\,\dee\tau\right) }_{\mbf{B}_{\mathrm{d}}}
\,\mbf{x}_{k},
\end{equation}
where $\mbf{z}_k = \mbf{z}(t_k)$ and $\mbf{x}_k = \mbf{x}(t_k)$, 

Let the DT Koopman representation of the VE dynamics be
\begin{equation}
\label{eqn: Koopman_evolvement}
    \mbs{\vartheta}_{k+1}=\mbf{A}\mbs{\vartheta}_k+\mbf{B}\mbf{u}_k,
\end{equation}
where $\mbs{\vartheta}_k \in \rnums^{p \times 1}$ is the vector that contains the lifted states, and $\mbf{A}$ and $\mbf{B}$ are defined as in (\ref{eqn: ss_koopman}). Rather than $\mbs{\upsilon}(\mbf{x}_k, \mbf{u}_k)$, only $\mbf{u}_k$ is assumed to be the input since the input usually appears linear in the dynamics and therefore does not need to be lifted in concert with $\mbf{x}_k$ \cite{korda2018linear}. Additionally, it will be shown in the case study in Section \ref{sec: results} that it is sufficient to lift only the states for our application. Since the coupling force depends on the difference between the device states and the VE states, and the VE is the black box that is learned, the coupling force can be treated as an exogenous input.

Since the input $\mbf{u}_k$ considered here for the VE is the coupling force, (\ref{eqn: Koopman_evolvement}) becomes
\begin{equation}
    \mbs{\vartheta}_{k+1}=\mbf{A}\mbs{\vartheta}_k+\mbf{B}(\mbf{M}\mbf{z}_k+\mbf{N}\mbf{x}_k),
\end{equation}
where $\mbf{M}=\bbm k_{\mathrm{c}} & b_{\mathrm{c}} \ebm$ and $\mbf{N}=\bbm -k_{\mathrm{c}} & -b_{\mathrm{c}} \ebm$. However, $\mbs{\vartheta}_k$ is defined such that the first $m$ terms of $\mbs{\vartheta}_k$ are $\mbf{x}_k$. Therefore, using $\mbftilde{N}=\bbm \mbf{N} & \mbf{0} \ebm$ where $\mbf{0} \in \rnums^{1 \times (p-m)}$ gives 
\begin{equation}
    \label{eqn: final_VE}
    \mbs{\vartheta}_{k+1}=(\mbf{A}+\mbf{B}\mbftilde{N})\mbs{\vartheta}_{k}+\mbf{B}\mbf{M}\mbf{z}_k.
\end{equation}

To construct the CL dynamics, (\ref{eqn: ss_device_DT}) also needs to be reformulated as 
\begin{equation}
    \label{eqn: final_device}
\mbf{z}_{k+1}=\mbf{A}_{\mathrm{d}}\mbf{z}_{k}+\mbf{B}_{\mathrm{d}}\mbf{S}\mbs{\vartheta}_k,
\end{equation}
where $\mbf{S}=\bbm \eye_{k} & \mbf{0}_{k \times (p-k)} \ebm$ and $k$ is the number of states in $\mbf{z}_{k}$.
Finally, combining (\ref{eqn: final_VE}) and (\ref{eqn: final_device}), the CL states are $\mbf{x}_{\mathrm{CL}}= \bbm \mbf{z}_k^\trans & \mbs{\vartheta}_k^\trans \ebm^\trans$, and the CL dynamics are 
\begin{equation}
\label{eqn: CL}
    \underbrace{\begin{bmatrix}
\mbf{z}_{k+1} \\
\mbs{\vartheta}_{k+1}
\end{bmatrix}}_{\mbf{x}_{\mathrm{CL}, k+1}}
=
\underbrace{\begin{bmatrix}
\mbf{A}_{\mathrm{d}} & \mbf{B}_{\mathrm{d}} \mbf{S} \\
\mbf{B} \mbf{M} & \left(\mbf{A} + \mbf{B} \mbftilde{N} \right)
\end{bmatrix}}_{\mbf{A}_{\mathrm{CL}}}
\underbrace{\begin{bmatrix}
\mbf{z}_{k} \\
\mbs{\vartheta}_{k}
\end{bmatrix}}_{\mbf{x}_{\mathrm{CL},k}}.
\end{equation}
Assessing CL stability amounts to a simple eigenvalue check on $\mbf{A}_{\mathrm{CL}}$. Specifically, CL asymptotic stability requires all eigenvalues to have magnitude strictly less than 1. In addition, designers can simply select the virtual coupling parameters, the sampling time step, or the device parameters subject to the constraint that all eigenvalues of $\mbf{A}_{\mathrm{CL}}$ have magnitude strictly less than 1. Moreover, the eigenvalue check on $\mbf{A}_{\mathrm{CL}}$ is less conservative than any passivity-based stability criterion, since passivity provides only a sufficient condition for stability, while the proposed eigenvalue check enables the stability analysis of a much broader class of systems, including those that are not passive \cite{khalil2002nonlinear}.


\subsection{Lyapunov's Direct Method}
Lyapunov's direct method for DT systems is a widely used method for stability analysis \cite{khalil2002nonlinear, bof2018lyapunov}. A suitable Lyapunov candidate is the total energy stored in the device, the virtual coupling, and the VE.

For the case study in this paper, the VE is considered to be a 1-DoF Duffing oscillator with dynamics
\begin{equation}
\label{eqn: Duffing_oscillator}
     m_{\mathrm{v}}\ddot{x_{\mathrm{v}}}(t) + b_{\mathrm{v}}\dot{x_{\mathrm{v}}}(t) + k_1 x_{\mathrm{v}}(t) + k_2 x_{\mathrm{v}}^3(t) = -F_{\mathrm{c}}(t),
\end{equation}
where the subscript ``v'' indicates variables and parameters for the VE, $m_{\mathrm{v}}$ is the mass, $b_{\mathrm{v}}$ is the damping coefficient, $k_1$ is the linear stiffness, $k_2$ is the nonlinear stiffness, and $F_{\mathrm{c}}(t)$ is the virtual coupling force from (\ref{eqn: device_dynamics}). Considering this nonlinear model for the VE, with the device dynamics in (\ref{eqn: device_dynamics}), the total energy stored in the entire system in DT is
\begin{equation}
\label{eqn: V_nonlinear}
\begin{aligned}
V_{\mathrm{nonlinear}}(\mbf{z}_{k}, \mbf{x}_{k})
&= \frac{1}{2} m_{\mathrm{d}}\dot{x}_{\mathrm{d},k}^{2} + \frac{1}{2} k_{\mathrm{c}} (x_{\mathrm{v},k} - x_{\mathrm{d},k})^{2}  \\
&\quad +\frac{1}{2} m_{\mathrm{v}} \dot{x}_{\mathrm{v},k}^{2} + \frac{1}{2} k_1 x_{\mathrm{v},k}^{2} + \frac{1}{4} k_2 x_{\mathrm{v},k}^{4},
\end{aligned}
\end{equation}
where $\mbf{z}_k$ and $\mbf{x}_k$ are defined as in (\ref{eqn: ss_device_DT}).
The Lyapunov candidate will be evaluated at each time step.

Similarly, the Lyapunov candidate for the haptic system with the Koopman-model VE is
\begin{equation}
\label{eqn: V_Koopman}
    V_{\mathrm{Koopman}}(\mbf{z}_k,\mbs{\vartheta}_k)=m_{\mathrm{d}} \dot{x}_{\mathrm{d},k}^2+k_{\mathrm{c}} (x_{\mathrm{v},k} - x_{\mathrm{d},k})^2+\mbs{\vartheta}_k^\trans\mbf{P}\mbs{\vartheta}_k,
\end{equation}
where $\mbf{P}=\mbf{P}^\trans \succ \mbf{0}$ satisfies $\mbf{A}^\trans \mbf{P} \mbf{A} - \mbf{P} = -\mbf{Q}$ with $\mbf{Q}=\mbf{Q}^\trans \succ \mbf{0}$ and $\mbf{A}$ being defined in (\ref{eqn: ss_koopman}). Although the Koopman-model VE allows a CL stability analysis via $\mbf{A}_{\mathrm{CL}}$ in (\ref{eqn: CL}), the Lyapunov candidate is included here for a comparison between the two methods.

%% file: Sections/Experiment.tex
\section{Experiment}
This section introduces validation experiments involving both the nonlinear-model VE and the Koopman-model VE, where the nonlinear model serves as the baseline for evaluating the Koopman representation. The corresponding simulations were also performed for comparison.

\subsection{Simulations}
Simulations of the haptic response using both the nonlinear-model VE and the Koopman-model VE were performed assuming the device EE dynamics in (\ref{eqn: device_dynamics}). The haptic device used in this study is introduced in Section \ref{sec: experimental_setup}. The effective mass $m_{\mathrm{d}}$ at the device EE is $0.287 \; \si{kg}$ \cite{kovacs2015coupled_dynamics}. The damping coefficient was tuned to match the baseline simulation with the baseline experimental results. The device and VE dynamics were updated using the fourth-order Runge–Kutta (RK4) method, with the simulations initialized at a nonzero position and a zero velocity, and evolved without external force input.

\subsection{Experimental Setup}
\label{sec: experimental_setup}
\begin{figure}[h]
    \centering
    \includegraphics[width=1\linewidth]{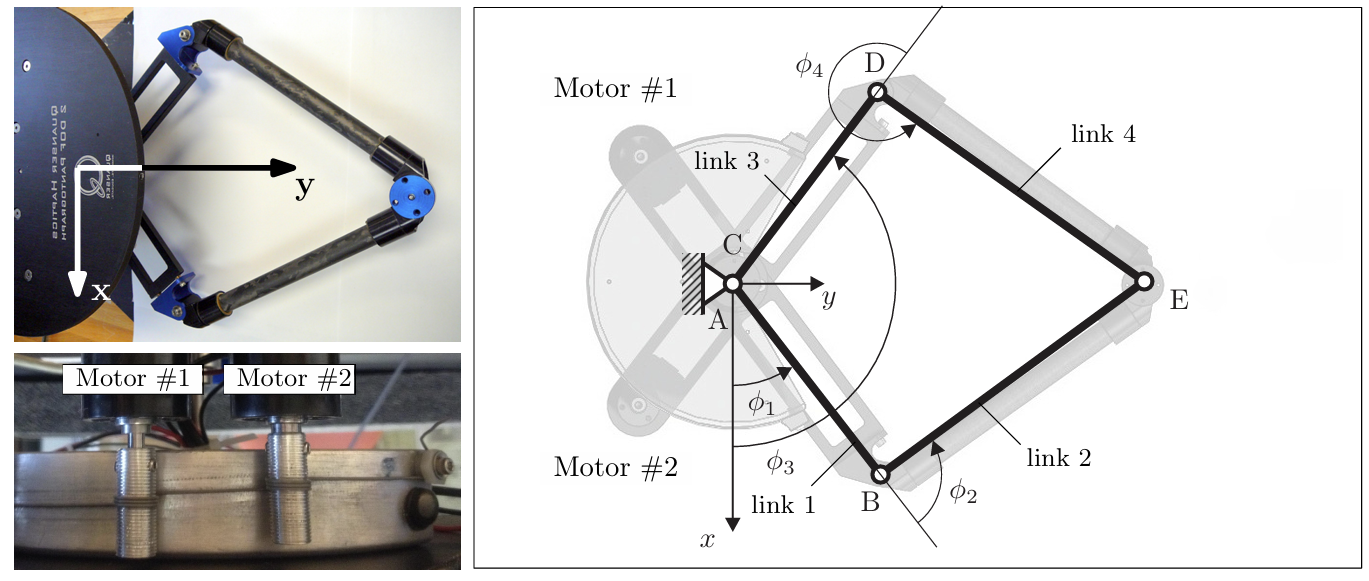}
    \caption{Quanser 2-DOF pantograph \cite{Sara_thesis_2014}.}
    \label{fig: pantograph}
\end{figure}

The experiments were performed on the Quanser 2-DOF Pantograph shown in Figure~\ref{fig: pantograph}. It is a five-bar linkage with two actuated DoFs driven by two DC motors (Maxon RE-26-1187742) equipped with quadrature encoders with 1000 counts per revolution. The pantograph is driven by a capstan drive mechanism. Quanser's hardward-in-the-loop (HIL) software development kit (SDK) is used to command the Quanser QPID board and the Quanser AMPAQ L2 amplifier that amplifies the control signal to drive the motors. 

A custom application to simulate the Duffing-oscillator VE using Quanser's HIL SDK was developed in C. The coupling force transmitted to the device EE in (\ref{eqn: coupling_force}) was calculated based on the readings on the device EE position and velocity. The VE position and velocity were updated corresponding to the VE dynamics. The EE positions were calculated through direct kinematics using the readings from the encoders. The EE velocities were obtained through differential kinematics (the Jacobian) using velocity counts that were computed in the QPID board from the encoder counts. The torque required at each joint was then obtained using the command force at the device EE and the Jacobian of the pantograph. The command current at each motor was then calculated using the motor torque constant and the capstan drive gear ratio. The sampling rate on the device and the integration rate for the VE were both $1000 \; \si{Hz}$.

The system parameters are summarized in Table \ref{table: system_parameters} \cite{quanserPantographManual}. In this table, the numerical subscripts correspond to the link number in Fig. \ref{fig: pantograph}; $m$ denotes the mass, and $l$ denotes the length.

\begin{table}[h]
\caption{System Parameters}
\label{table: system_parameters}
\begin{center}
\begin{tabular}{|c|c|}
\hline
$m_1$, $m_3$ & $0.282 \; \si{kg}$ \\
\hline
$m_2$, $m_4$ & $0.05427 \; \si{kg}$\\
\hline
$l_1$, $l_3$ & $0.147 \; \si{m}$\\  
\hline
$l_2$, $l_4$ & $0.199 \; \si{m}$\\
\hline
Motor torque constant & $0.0502 \; \si{Nm/A}$\\
\hline
Capstan gear ratio & $19.4$\\
\hline
\end{tabular}
\end{center}
\end{table}


\subsection{Single-User Study}
A Duffing-oscillator VE with the nonlinear model in (\ref{eqn: Duffing_oscillator}) with $m_{\mathrm{v}}=1\;\si{kg}$, $b_{\mathrm{v}}=5 \; \si{Ns/m}$, $k_1=100 \; \si{N/m}$, $k_2=-3 \; \si{N/m^3}$ was examined in experiments as the baseline. The VE moves in the $y$-direction as indicated in Fig. \ref{fig: pantograph}. The virtual coupling parameters were chosen to be $k_{\mathrm{c}}=4000 \; \si{N/m}$ and $b_{\mathrm{c}}=8 \; \si{N/m}$. The integration of the baseline VE follows the RK4 method as in the simulation. A Koopman model was trained using \texttt{pykoop} \cite{pykoop} to compare with the baseline model in simulation and experiment.

To obtain the Koopman model, $25$ datasets were simulated from the nonlinear model (\ref{eqn: Duffing_oscillator}) using inputs of sine and cosine waves with various amplitudes ranging from $0.1$ to $4.0$ and different frequencies ranging from $\f{1}{2\pi} \; \si{Hz}$ to $20 \; \si{Hz}$, starting from initial positions $\bbm x_{\mathrm{v}0} & \dot{x}_{\mathrm{v}0} \ebm$ where $x_{\mathrm{v}0} \in \bbm -1, \;1\ebm \si{m}$ and $\dot{x}_{\mathrm{v}0} \in \bbm -1, \;0.5\ebm \si{m/s}$. The inputs were selected according to the form of the virtual coupling force in the simulation. The initial conditions were selected to cover values for the experimental application and the workspace of the pantograph, with additional out-of-range cases for generalization. Since the nonlinear dynamics contain a cubic dependency in the position, the lifting functions were chosen to be polynomials up to order $3$, namely
$\mbs{\vartheta}(x_{\mathrm{v}}, \dot{x}_{\mathrm{v}})=\bbm x_{\mathrm{v}} & \dot{x}_{\mathrm{v}} & x_{\mathrm{v}}^2 & \dot{x}_{\mathrm{v}}^2 & x_{\mathrm{v}}\dot{x}_{\mathrm{v}} & x_{\mathrm{v}}^3 & x_{\mathrm{v}}^2\dot{x}_{\mathrm{v}} & x_{\mathrm{v}}\dot{x}_{\mathrm{v}}^2 & \dot{x}_{\mathrm{v}}^3\ebm^\trans$ where the argument $t_k$ has been omitted. It is intuitive to first consider state-dependent lifting functions since the input appears linearly. In the Koopman update via (\ref{eqn: Koopman_evolvement}), the states were retracted and relifted at each step to prevent error propagation in (\ref{eqn: ss_koopman}) \cite{mamakoukas2020learning}.

The Duffing-oscillator VE was located at $y=0.2 \; \si{m}$. Experiments were conducted by moving the EE of the pantograph towards the positive $y$-direction. Once the virtual Duffing oscillator was reached, it was further displaced by $0.075 \; \si{m}$ and subsequently released. This procedure resulted in the virtual Duffing oscillator responding from an initial position under free force input. Experiments were carried out for both the nonlinear-model VE and the Koopman-model VE.

\subsection{Multi-User Study}
To assess whether users perceive similar force feedback from the Koopman-model VE and the baseline nonlinear-model VE, a multi-user study was conducted. The users gave their consent to the anonymous reporting of their responses for academic use.

Participants were instructed to perceive the Duffing-oscillator VEs by alternatively pulling and pushing the mass between two designated positions, $y=0.29 \; \si{m}$ and $y=0.17 \; \si{m}$. These target locations were specifically indicated using reference objects. Each participant repeated the same motion at approximately the same speed three times for both VEs and was asked to recall their haptic impressions. Since haptic perception is subjective, to obtain reliable results, all participants experienced the baseline VE twice and the Koopman-model VE once. Only participants who reported perceiving consistent force feedback for the two baseline trials were considered reliable for the subsequent analysis.

Participants rated, on a scale ranging from $0$ (strongly disagree) to $10$ (strongly agree), the extend to which they perceived the nonlinear-model VE and the Koopman-model VE to be closely aligned. 
To determine whether the perceived similarity was significantly above the neutral agreement, a one-sample Wilcoxon signed-rank test \cite{woolson2007wilcoxon} was conducted. The test evaluated the null hypothesis ($H_0$) that the median similarity rating was less than or equal to $6$ (threshold of agreement) against the alternative hypothesis ($H_a$) that the median rating was greater than $6$.


%% file: Sections/results.tex
\section{Results and Discussions}
\label{sec: results}
\subsection{Simulations}
\begin{figure}[h]
    \centering
    \includegraphics[width=1\linewidth]{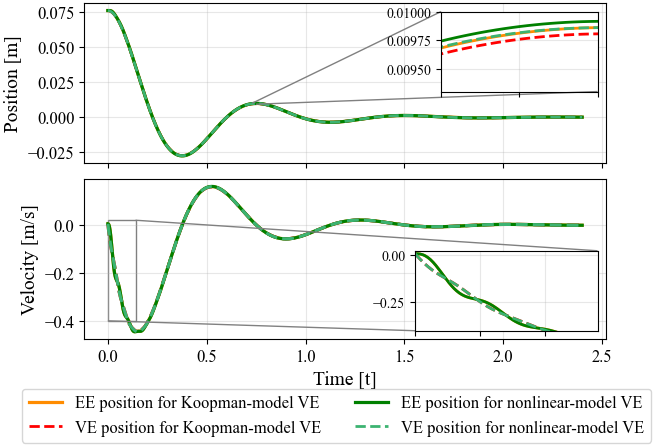}
    \caption{Simulated responses for nonlinear-model and Koopman-model VEs.}
    \label{fig:simulation}
\end{figure}

The simulated device EE and VE responses for both the nonlinear-model VE and the Koopman VE are summarized in Fig. \ref{fig:simulation}. The ICs were assumed to be an ideal case with $x_{d0}=x_{v0}=0.075\;\si{m}$ and $\dot{x}_{d0}=\dot{x}_{v0}=0 \;\si{m/s}$.

The solid green curves show the device EE states for the nonlinear model of the Duffing-oscillator VE. The dotted light-green curves depict the VE states for the nonlinear-model VE. The solid orange curves represent the device EE states when using a Koopman model of the Duffing-oscillator VE. The dotted red curves denote the VE states for the Koopman-model VE. The simulated responses of both the position and the velocity align closely between the Koopman and nonlinear models, demonstrating that the Koopman representation effectively reproduces the dynamics of the nonlinear-model VE. 

A mismatch appears between the device EE and VE velocities at the beginning of the simulation. This occurs because the device acceleration is $0 \;\si{m/s^2}$ in (\ref{eqn: device_dynamics}), where structural stiffness is neglected. In contrast, the VE has a nonzero acceleration determined by the initial position and stiffness coefficients. Therefore, the initial velocities do not align. This mismatch is subsequently resolved by the virtual coupling in the following manner. The coupling force (\ref{eqn: coupling_force}) pulls the device EE towards the positive $y$-direction and pushes the VE backward when it is positive, and pulls the VE forward and pushes the device EE backward when it is negative. A larger virtual damping in the coupling smooths out the initial velocity mismatch but reduces interaction rigidity between the haptic device and the VE. The virtual-coupling parameters were chosen to ensure sufficient smoothness while preserving an acceptable level of user experience.

\subsection{Single-User Study}
\begin{figure}[h]
    \centering
    \includegraphics[width=1\linewidth]{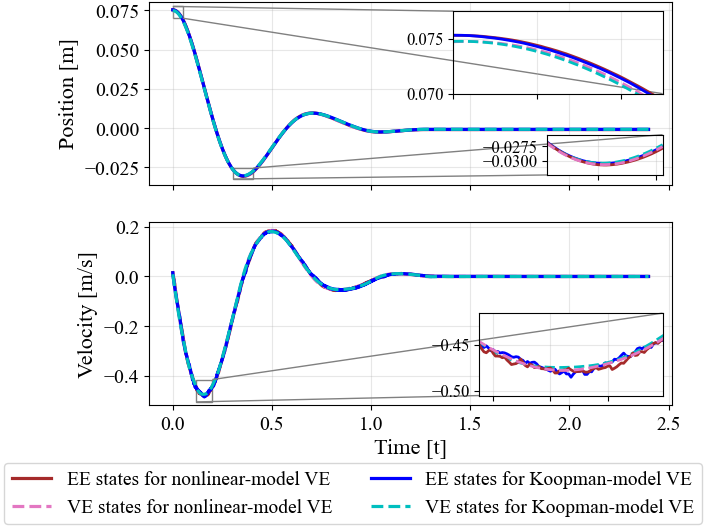}
    \caption{Experimental responses for nonlinear-model and Koopman-model VEs.}
    \label{fig:Exp_results}
\end{figure}

The experimental results are summarized in Figure~\ref{fig:Exp_results}. Due to human error, the initial conditions (ICs) of the two trials are not identical, but are close enough for a comparison. Table \ref{table: initial_conditions} summarizes the ICs.

\begin{table}[h]
\centering
\caption{Initial Conditions Associated with Both VEs}
\begin{tabular}{c|c|c|c|c}
    \hline
     & $x_{d0}\,[\si{m}]$ & $x_{v0}\,[\si{m}]$ & $\dot{x}_{d0}\,[\si{m/s}]$ & $\dot{x}_{v0}\,[\si{m/s}]$ \\
    \hline
    \textbf{Nonlinear model} & 0.07531 & 0.07472 & 0.01500 & 0.01762 \\
    \hline
    \textbf{Koopman model} & 0.07532 & 0.07476 & 0.01490 & 0.00885 \\
    \hline
\end{tabular}
\label{table: initial_conditions}
\end{table}
According to Figure~\ref{fig:Exp_results}, the responses for the two VEs are almost identical. Furthermore, the alignment between the VE states and the device EE states in both cases demonstrates that both VEs achieve a satisfying level of fidelity. The velocity curves are less smooth than the position curves due to noise amplification and quantization effects in numerical differentiation. Overall, the experimental results confirm that the Koopman-model VE delivers a valid representation of the Duffing-oscillator VE.

\begin{figure}
    \centering
    \includegraphics[width=1\linewidth]{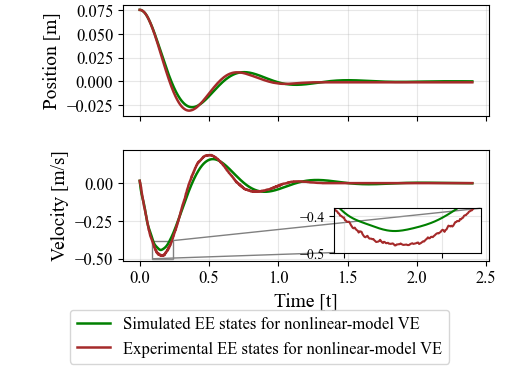}
    \caption{Experimental vs. simulation EE states for nonlinear-model VE.}
    \label{fig:Exp_vs_Sim_nl}
\end{figure}

\begin{figure}[h]
    \centering
    \includegraphics[width=1\linewidth]{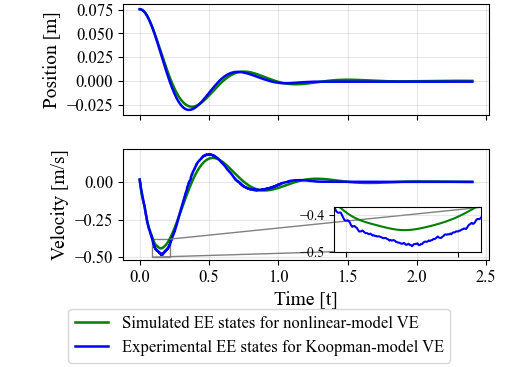}
    \caption{Experimental vs. simulation EE states for Koopman-model VE.}
    \label{fig:Exp_vs_Sim_kp}
\end{figure}

Since the two experiments have slightly different ICs, for a more detailed comparison, the simulated baseline responses corresponding to these ICs were compared to the experimental responses for the nonlinear-model VE in Figure~\ref{fig:Exp_vs_Sim_nl} and to those for the Koopman-model VE in Figure~\ref{fig:Exp_vs_Sim_kp}. Since Figure~\ref{fig:Exp_results} shows a close alignment between the VE and device EE responses, the comparisons in Figure~\ref{fig:Exp_vs_Sim_nl} and Figure~\ref{fig:Exp_vs_Sim_kp} consider only the states of the device EE.


The experimental responses do not coincide exactly with simulation. Although the device damping in (\ref{eqn: device_dynamics}) was tuned to minimize the root mean square error (RMSE) between the simulated and experimental baseline result, (\ref{eqn: device_dynamics}) is a simplified representation for the device dynamics, neglecting structural flexibility from the joints and the capstan drive. Therefore, for both the VEs, the experimental responses exhibit a softer behaviour than the simulation. 

Although the primary goal is not to achieve a close match between simulation and experiment, comparing the two provides meaningful insights. Table \ref{table: RMSE} summarizes the RMSE comparing the simulated baseline responses and the experimental results obtained using the two VEs. To mitigate noise and quantization effects in the RMSE computation, the experimental data was preprocessed by a third-order Butterworth low-pass filter with a cut-off frequency of $50 \; \si{Hz}$.

\begin{table}[h]
\centering
\caption{RMSE Between the Responses from the Simulated Baseline and Experiments}
\begin{tabular}{c|c|c}
    \hline
     & \textbf{Nonlinear-model VE} &  \textbf{Koopman-model VE} \\
    \hline
     RMSE in $x_{\mathrm{d}} \;\left[\si{m}\right]$ & $0.002214$ & $0.002038$ \\
    \hline
     RMSE in $\dot{x}_d \;\left[\si{m/s}\right]$ & $0.02003$  & $0.01872$  \\
    \hline
\end{tabular}
\label{table: RMSE}
\end{table}

The RMSE values for the Koopman-model VE are lower than those for the nonlinear-model VE, indicating that the responses associated with the Koopman-model VE are closer to the ideal simulation that assumes a rigid-body haptic device.
Since the dominant simulation-experiment discrepancy is caused by the sturctural flexibiliy in the haptic device, these results suggest that the Koopman-model VE is more robust to structural uncertainties of the device. This can be explained as follows. 

Due to structural flexibility, the measured EE position is different from the ideal $x_{\mathrm{d}}(t)$ under the rigid-body-device assumption. This inherently leads to a deviation in the actual VE position compared to the ideal VE position $x_{\mathrm{v}}(t)$, since the virtual coupling force aims to align the VE position with the device EE position. Let the experimental VE position be $x_{\mathrm{v},\mathrm{exp}}(t)=x_{\mathrm{v}}(t)+\delta_{\mathrm{v}}(t)$, where $\delta_{\mathrm{v}}(t)$ is the perturbation in the VE position. In the nonlinear-model VE, this perturbation enters the cubic term, which amplifies the small error. However, in the Koopman representation, the dynamics are linear in the lifted space. Although the term $x_{\mathrm{v}}^3(t)$ is involved in the lifted states, the Koopman identification process learns a model that best represents the dynamics in terms of a linear combination of all the lifted states. Therefore, it does not assign an outsized importance to the $x_{\mathrm{v}}^3(t)$ term alone. This makes the overall model more robust to small deviations in the state variables compared to the baseline model. 



In addition, to confirm that this advantage of using the Koopman-model VE is not a coincidence, the RMSE values for $9$ additional groups of experiments with different ICs were computed and summarized in Table \ref{tab:init-rmse} in the appendix. The results demonstrate that the Koopman-model VE consistently produces lower RMSE values than the nonlinear-model VE. Additionally, it is observed that, for the same model, the larger the initial positions for the device EE and the VE, the larger the RMSE. This can also be explained by the sensitivity to perturbations in $x_{\mathrm{v}}$. Since the error introduced by the perturbation $\delta_{\mathrm{v}}$ is associated with $x_{\mathrm{v}}^3\delta_{\mathrm{v}}$ (but with different weightings) for both models, a larger $x_{\mathrm{v}}$ leads to a larger RMSE compared to the simulation baseline. The consistency of the results is further confirmed by the fact that the Koopman-model VE maintains lower RMSE values even in trials where its initial positions were larger than those of the nonlinear-model VE.

\subsection{Stability Analysis}
As discussed in Section \ref{sec: Closed-Loop Stability Analysis}, using the Koopman model to represent the nonlinear VE dynamics provides a more convenient CL stability analysis. 
The results on evaluating the eigenvalues of $\mbf{A}_{\mathrm{CL}}$ in (\ref{eqn: CL}) and the difference of the Lyapunov candidates in (\ref{eqn: V_nonlinear}) and (\ref{eqn: V_Koopman}) at each time step confirms this advantage. 

\begin{figure}
    \centering
    \includegraphics[width=1\linewidth]{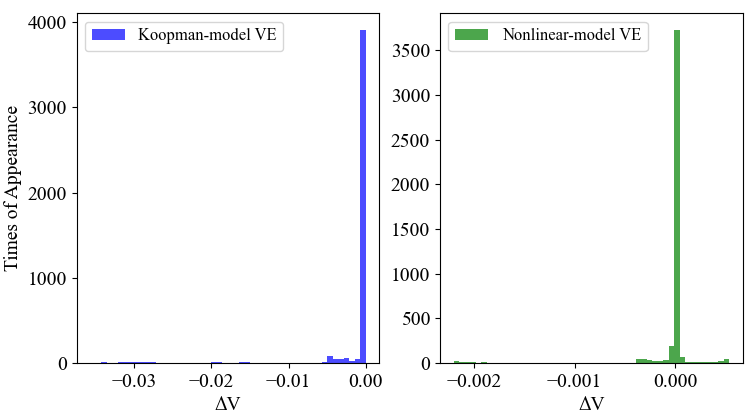}
    \caption{Histogram of Lyapunov candidate difference for Koopman-model VE (left) and nonlinear-model VE (right).}
    \label{fig:histgram}
\end{figure}

Using experimental data, the maximum per-step difference in the Lyapunov candidate is $5.361\times 10^{-4}$ for $V_\mathrm{nonlinear}$ associated with the baseline VE and $2.376\times 10^{-6}$ for $V_\mathrm{Koopman}$ associated with the Koopman-model VE. Neither value is strictly negative.
As shown in Figure~\ref{fig:histgram}, the difference in the Lyapunov candidate, $\Delta V$, for the Koopman-model VE is negative or closed to $0$ the majority of the time, while that for the nonlinear-model VE contains some values greater than $0$.
 
\begin{figure}
    \centering
    \includegraphics[width=0.85\linewidth]{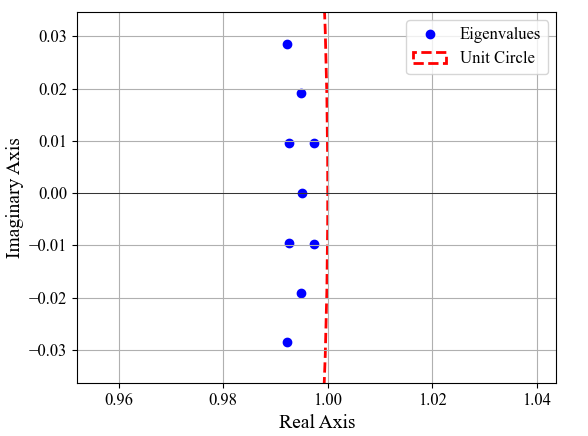}
    \caption{Eigenvalues of $\mbf{A}_{\mathrm{CL}}$.}
    \label{fig:eigenvalues}
\end{figure}

However, the aforementioned results involving $V_\mathrm{nonlinear}$ and $V_\mathrm{Koopman}$ are inconclusive. Assessing either Lyapunov or asymptotic stability of the haptic system associated with the nonlinear-model VE or the Koopman-model VE would require finding alternative Lyapunov candidates, which is challenging. By contrast, the CL analysis with the Koopman-model VE in the case study shows that all eigenvalues of $\mbf{A}_{\mathrm{CL}}$ in (\ref{eqn: CL}) lie strictly inside the unit circle as shown in Figure~\ref{fig:eigenvalues}, indicating asymptotic stability. 

Overall, although Lyapunov's direct method is applicable for both VEs, constructing a definitive Lyapunov candidate is difficult. With the Koopman-model VE, the CL stability analysis avoids this challenge. This advantage makes the stability analysis of the CL haptic system more accessible than the Lyapunov's direct method and less conservative compared to passivity-based stability criteria, while also simplifying the parameter-design process to ensure CL stability.

However, the Koopman model presented here for the Duffing-oscillator VE is not a unique representation. A Koopman model using $8$ lifting functions from the set of third-order monomials excluding $x_{\mathrm{v}}^3(t)$ also provides a reasonable theoretical match, but yields a noticeably higher RMSE than the one with $9$ lifting functions. This highlights the sensitivity to the choice of lifting functions. Despite this modeling error, the CL system remains asymptotically stable. However, selecting an appropriate set of lifting functions remains a practical challenge due to the absence of a systematic design methodology \cite{Dahdah_2022_sysNorm_regularization, mezic2023operator, colbrook2024limits}. 

\subsection{Multi-User Study}
Table \ref{table: user_study} summarizes the rating from reliable users who participated in the multi-user study. 

\begin{table}[h]
\caption{User Study Results}
\label{table: user_study}
\begin{center}
\begin{tabular}{c|ccccccccccc}
\hline
\textbf{User} & 1 & 2 & 3 & 4 & 5 & 6 & 7 & 8 & 9 & 10 & 11 \\
\hline
\textbf{Rating} & 9 & 8 & 8 & 10 & 10 & 6 & 8 & 9 & 5 & 9 & 10 \\
\hline
\end{tabular}
\end{center}
\end{table}

The mean of the ratings on the similarity of the two VEs is $8.364$, and the median is $9$, indicating that the users generally perceived a high degree of similarity between the two VEs. The one-sample Wilcoxon signed-rank test further confirmed that the median rating is statistically significantly greater than $6$ ($p<0.004$), the threshold for agreement on the scale. This provides evidence that user agreement was not marginal but consistent, supporting the conclusion that the Koopman model offers an effective representation of the nonlinear Duffing oscillator dynamics. 


\subsection{Computational Complexity}
Although the Koopman-model VE has several advantages, the computation for each time step is more complex than using an RK4 method to update a nonlinear model. For a nonlinear model with $k$ states, the computational complexity for RK4 is $\mc{O}(k)$. For the Koopman-model VE, since relifting and retracting with a vector-value lifting function in the form of (\ref{eqn: vector-valued lifting function}) is not computationally complex, the main complexity comes from the matrix multiplication to advance the lifted state vector, which is $\mc{O}(p^2)$, where $p$ is the dimension of the lifted-state vector. Although the Koopman-model VE advancement is more computationally complex than the nonlinear-model VE update with RK4, they both have polynomial complexities. The value of $p$ for a Koopman model is usually modest, thus the cost remains manageable. By contrast, if the nonlinear-model VE integration requires an implicit RK method for a numerically stable update, the complexity becomes $\mc{O}(k^3)$. In such cases, the Koopman-model VE update becomes more efficient.  


%% file: Sections/conclusions.tex
\section{Conclusions}
This paper introduced and investigated the novel use of the Koopman operator to represent nonlinear VEs for kinesthetic haptic rendering and couple it to the human operator. The Koopman representation of nonlinear VEs enables the use of well-established linear tools to the analysis of nonlinear systems. In a case study of simulating a Duffing-oscillator VE, the Koopman model provides an effective representation of the nonlinear dynamics. The Koopman-model VE provides a more convenient CL stability analysis than the nonlinear-model VE since it is linear in the lifted space. Such stability criterion is also less conservative than passivity-based criteria since Lyapunov and asymptotic stability encompass a broader class of systems than passivity. In addition, the Koopman-model VE demonstrates more robustness to small deviations in the VE states than the nonlinear-model VE, indicating that the Koopman-model VE is less sensitive to device-modeling uncertainties when performing virtual-coupling parameter design or other control designs in simulation. Although the Koopman-model VE requires more complex computation compared to the RK4-method-updated nonlinear-model VE, performing matrix computation with $\mc{O}(p^2)$ is not very expensive. In contrast, if the nonlinear model in the VE requires implicit RK methods for numerical stability, using the Koopman-model VE is more computationally efficient. 


  

%% file: Sections/appendix.tex
\section*{Appendix}
\begin{table}[h]
\caption{RMSE Values for Koopman-model (KP) VE and nonlinear-model (NL) VE with different initial conditions}
\label{tab:init-rmse}
\centering
\scriptsize
\setlength{\tabcolsep}{3.5pt} 
\renewcommand{\arraystretch}{1.05} 
\resizebox{\linewidth}{!}{%
\begin{tabular}{c c c c c c c c}
\hline
Trial & Model & $x_{\mathrm{d}0}\;[\si{m}]$ & $\dot{x}_{\mathrm{d}0}\;[\si{m/s}]$ & $x_{\mathrm{v}0}\;[\si{m}]$ & $\dot{x}_{\mathrm{v}0}\;[\si{m/s}]$ & $\mathrm{RMSE}_{x}\;[\si{m}]$ & $\mathrm{RMSE}_{\dot x}\;[\si{m/s}]$ \\
\hline
1 & KP & $0.07726$ & $0.07665$ & $0.01211$ & $0.02197$ & $0.002154$ & $0.01957$ \\
1 & NL & $0.07600$ & $0.07548$ & $0.01764$ & $0.01921$ & $0.002259$ & $0.02054$ \\
\hline
2 & KP & $0.08707$ & $-0.0003390$ & $0.08585$ & $0.008988$ & $0.001861$ & $0.02147$ \\
2 & NL & $0.08606$ & $0.01394$ & $0.08573$ & $0.01029$ & $0.002691$ & $0.02396$ \\
\hline
3 & KP & $0.1050$ & $0.01491$ & $0.1039$ & $0.01223$ & $0.002278$ & $0.02600$ \\
3 & NL & $0.1081$ & $0.001320$ & $0.1069$ & $0.01325$ & $0.002409$ & $0.02611$ \\
\hline
4 & KP & $0.08846$ & $0.01723$ & $0.08776$ & $0.02085$ & $0.002277$ & $0.02137$ \\
4 & NL & $0.08646$ & $0.01714$ & $0.08609$ & $0.003763$ & $0.002557$ & $0.02311$ \\
\hline
5 & KP & $0.08544$ & $0.01042$ & $0.08442$ & $0.006680$ & $0.002023$ & $0.02160$ \\
5 & NL & $0.08587$ & $0.007006$ & $0.08547$ & $0.0005780$ & $0.002550$ & $0.02308$ \\
\hline
6 & KP & $0.07735$ & $0.01076$ & $0.07686$ & $-0.007429$ & $0.002156$ & $0.01980$ \\
6 & NL & $0.07601$ & $0.01764$ & $0.07548$ & $0.01921$ & $0.002259$ & $0.02054$ \\
\hline
7 & KP & $0.07092$ & $0.01685$ & $0.07025$ & $0.01408$ & $0.001579$ & $0.01616$ \\
7 & NL & $0.07181$ & $0.01889$ & $0.07119$ & $0.02082$ & $0.002184$ & $0.01923$ \\
\hline
8 & KP & $0.09659$ & $0.01872$ & $0.09584$ & $0.02418$ & $0.002536$ & $0.02497$ \\
8 & NL & $0.09313$ & $0.01227$ & $0.09266$ & $-0.0006020$ & $0.002718$ & $0.02551$ \\
\hline
9 & KP & $0.07329$ & $0.07239$ & $0.01035$ & $0.01280$ & $0.001685$ & $0.01768$ \\
9 & NL & $0.07606$ & $0.07514$ & $0.01527$ & $0.01997$ & $0.001797$ & $0.01855$ \\
\hline
\end{tabular}}%
\end{table}

%% file: Sections/acknowledgement.tex
\section*{Acknowledgment}
The authors would like to thank Dr. Steven Dahdah for their insights in the Koopman operator theory as well as the implementation of \texttt{pykoop} and Antoine Henri for their support on the development of the Quanser HIL SDK in C. This research was funded by McGill Engineering Doctoral Award, Fonds de recherche du Québec, Natural Sciences and Engineering Research Council of Canada (NSERC), and CM Labs Simulations, Inc. The support is gratefully acknowledged.